\documentclass[conference]{IEEEtran}
\IEEEoverridecommandlockouts
\usepackage{cite}
\usepackage{amsmath,amssymb,amsfonts}
\usepackage{algorithmic}
\usepackage{graphicx}
\usepackage{textcomp}

\usepackage{siunitx}
\usepackage{url}
\usepackage{multirow}
\usepackage{diagbox}
\usepackage{multicol, blindtext}
\usepackage{lipsum}

\usepackage{balance}

\usepackage{ amssymb }

\def\BibTeX{{\rm B\kern-.05em{\sc i\kern-.025em b}\kern-.08em
    T\kern-.1667em\lower.7ex\hbox{E}\kern-.125emX}}

\makeatletter
\def\ps@IEEEtitlepagestyle{%
  \def\@oddfoot{\mycopyrightnotice}%
  \def\@evenfoot{}%
}
\def\mycopyrightnotice{%
  {\footnotesize XXX-X-XXXX-XXXX-X/XX/\$xx.00 ©2019 IEEE\hfill}
  \gdef\mycopyrightnotice{}
}

\begin{document}

\title{End-to-end Cloud Segmentation in High-Resolution Multispectral Satellite Imagery Using Deep Learning}
\author{%
	\IEEEauthorblockN{%
		Giorgio Morales, 
		Alejandro Ram\'{i}rez,
		Joel Telles
		}
	\IEEEauthorblockA{%
		National Institute of Research and Training in Telecommunications (INICTEL-UNI)\\
		National University of Engineering, Lima, Peru\\
		Email: giorgiomoralesluna@gmail.com}%
}

\maketitle


\begin{abstract}
Segmenting clouds in high-resolution satellite images is an arduous and challenging task due to the many types of geographies and clouds a satellite can capture. Therefore, it needs to be automated and optimized, specially for those who regularly process great amounts of satellite images, such as governmental institutions. In that sense, the contribution of this work is twofold: We present the CloudPeru2 dataset, consisting of 22,400 images of $\mathbf{512\times512}$ pixels and their respective hand-drawn cloud masks, as well as the proposal of an end-to-end segmentation method for clouds using a Convolutional Neural Network (CNN) based on the Deeplab v3+ architecture. The results over the test set achieved an accuracy of 96.62\%, precision of 96.46\%, specificity of 98.53\%, and sensitivity of 96.72\% which is superior to the compared methods.\footnote{This paper is a preprint (submitted to the INTERCON 2019 conference, Lima, Peru). IEEE copyright notice. Ó 2019 IEEE. Personal use of this material is permitted. Permission from IEEE must be obtained for all other uses, in any current or future media, including reprinting/republishing this material for advertising or promotional purposes, creating new collective works, for resale or redistribution to servers or lists, or reuse of any copyrighted.}

\end{abstract}


\section{Introduction}

Since the very beginning of remote sensing, clouds represents the most overwhelming type of noise in optical satellite imagery because it blocks everything beneath them. On the other hand, the high variance in its spectral response could add statistical noise into a database if some of its pixels get into it. For those reasons, filtering clouds through a detection process is one of the most traditional problems in remote sensing. 

In the literature, this problem has been addressed from many perspectives; from empirical thresholded decision trees \cite{no_ml1,5}, fuzzy logic \cite{fuzzy}, time series (if data available) \cite{6}, and machine learning \cite{1,2,cloudperuann} to a more recent approach: deep learning \cite{4,9,cloudicann}. Even though some of the previous works achieved outstanding results, due the high risk that clouds represents, generating more accurate models for clouds detection is still valuable to enhance the results of deeper remote sensing methods/algorithms.

Due to the fact that some institutions, such as the National Commission for Aerospace Research and Development (CONIDA) of Peru, process a great number of satellite images daily, it is necessary to develop a method to automatically and rapidly obtain their correspondent cloud masks. For this, we propose an efficient cloud segmentation  method for high-resolution multispectral satellite images using a trainable end-to-end convolutional neural network (CNN). In order to train our network and compare its performance with other methods, we propose a large dataset consisting of 22,400 image patches extracted from PERUSAT-1, a Peruvian satellite managed and  supervised by CONIDA.

\section{Proposed Method}

\subsection{CloudPeru2 Dataset}

A PERUSAT-1 scene has four spectral bands: red (0.63-0.7\SI{}{\micro\metre}), green (0.53-0.59\SI{}{\micro\metre}), blue (0.45-0.50\SI{}{\micro\metre}) and NIR (0.752-0.885\SI{}{\micro\metre}). The spatial resolution of the multispectral bands is 2.8 m per pixel and that of the panchromatic band is 0.7 m per pixel. 

\begin{figure}
\centering
\includegraphics[height=4.4cm]{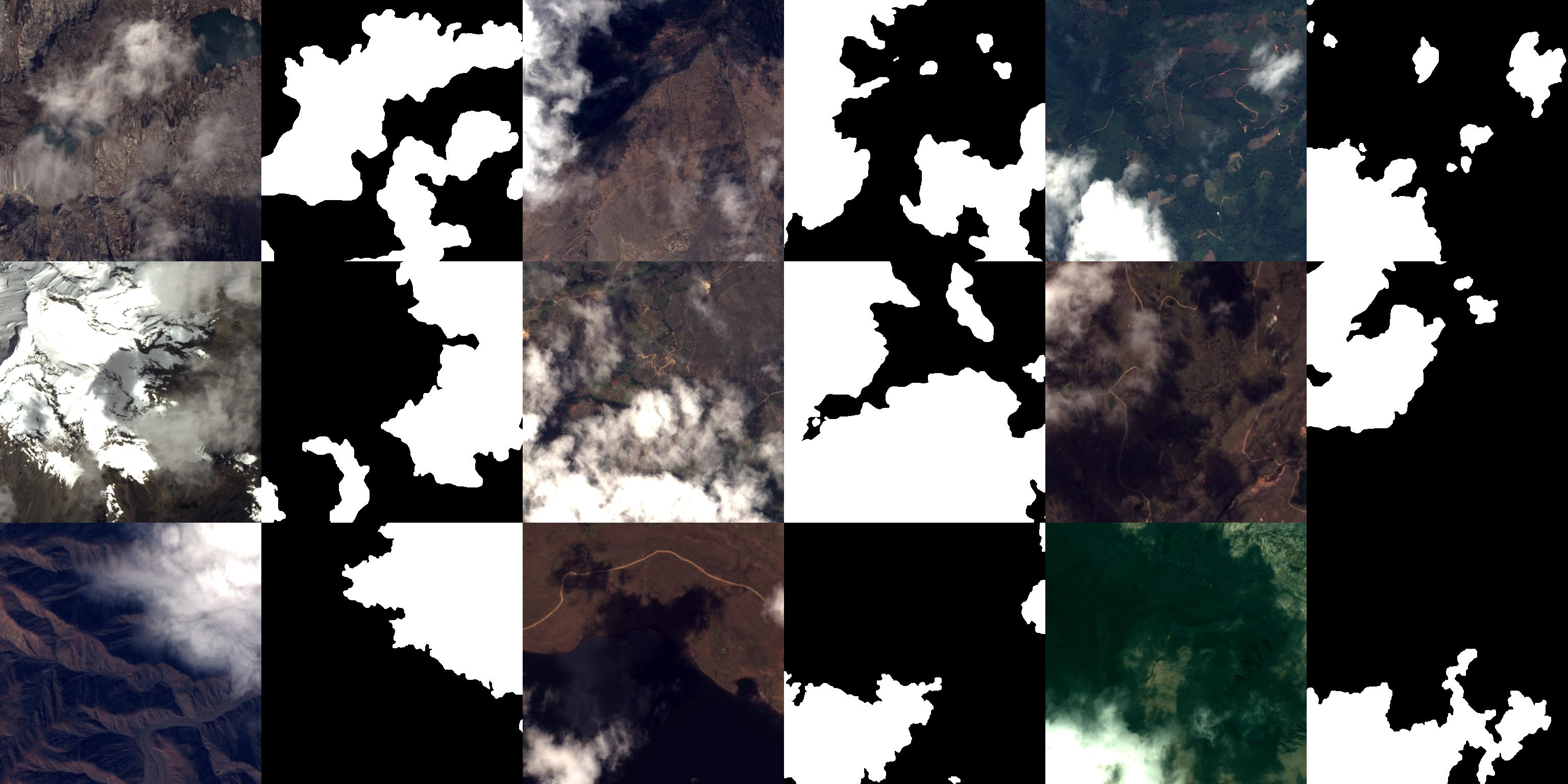}
\caption{Samples of original images and cloud masks from CloudPeru2 dataset.}
\label{fig:fig1}
\vspace{-2ex}
\end{figure}

We used 153 PERUSAT-1 scenes of variable sizes (from $6176\times 6012$ to $12722\times 9529$ pixels) and from different geographies to extract 2800 image patches of $512\times 512$ pixels and create the CloudPeru2 dataset \cite{cloudperu2}. The scenes were previously orthorectificated and adjusted to reflectance values with atmosferic correction. Each image patch has a correspondent hand-drawn shadow mask. Image samples from the dataset are shown in Fig.~\ref{fig:fig1}. Nevertheless, for this work we used data augmentation to increase the dataset size in order to avoid overfitting problems. In that sense, we rotated each patch \ang{90}, \ang{180} and \ang{270}, and flipped horizontally each one so that we get a total of 22,400 patches. We split 90\% of the data to create the training set, 5\% to the validation set and 5\% to the test set. 

In a previous work \cite{cloudicann}, we presented the CloudPeru dataset, which was used to classify small image patches as clouds or non-clouds. This dataset consists of 476,422 image patches of $27\times27$ pixels extracted from only 15 different PERUSAT-1 scenes. In contrast, the CloudPeru2 dataset presents a greater number of scenarios (e.g. snow and ocean) and a bigger patch size; besides, it was specifically created to solve a segmentation problem.   

\begin{figure}
\centering
\includegraphics[width=\columnwidth]{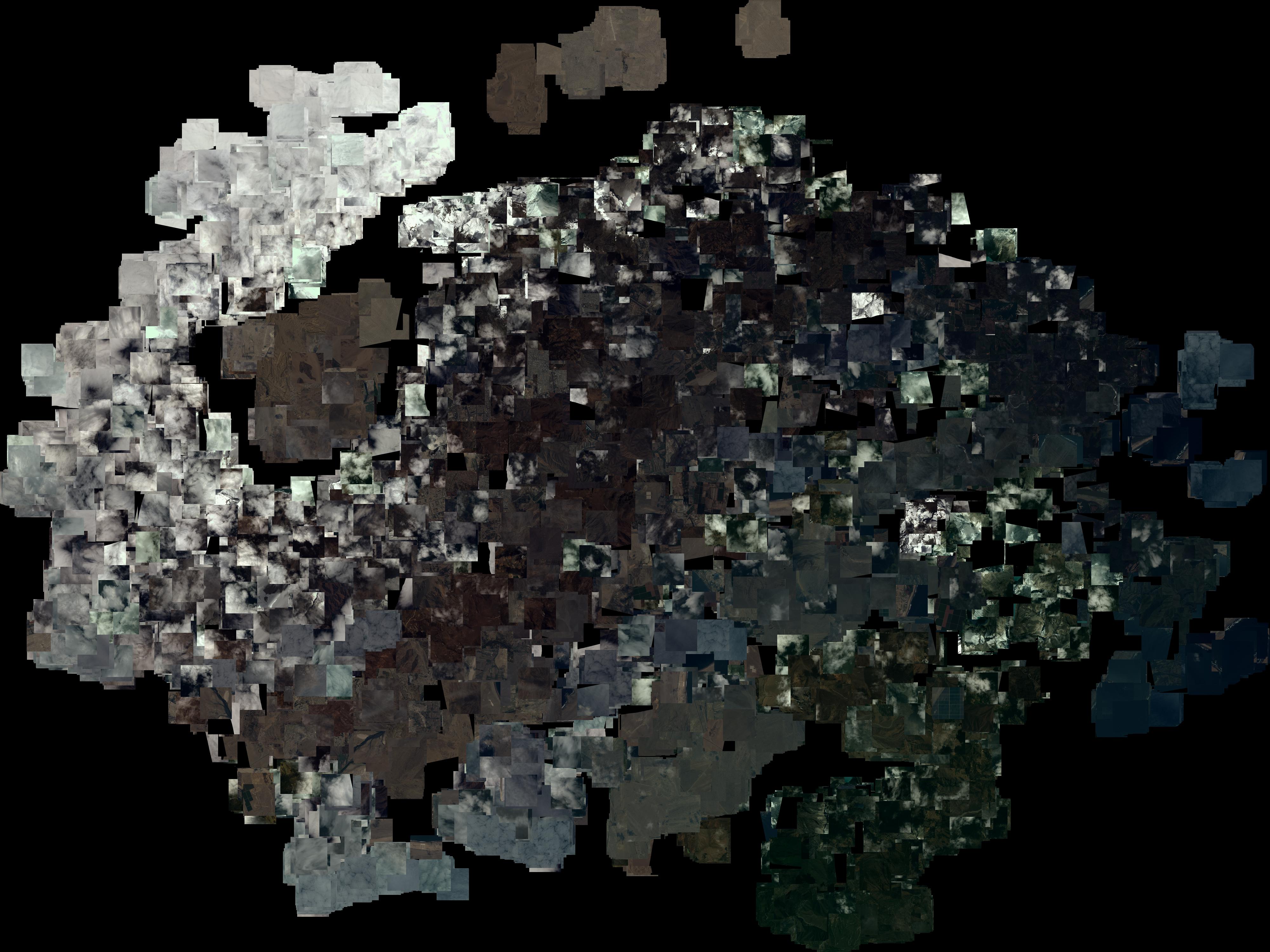}
\caption{Visualization of the CloudPeru2 dataset using t-SNE. Images fully covered by clouds are clustered in the left region. In the upper middle region are images of desert; bellow, snowy mountains and urban areas. Images of forest and ocean are clustered at the right.}
\label{fig:fig2}
\vspace{-2ex}
\end{figure}

In order to appreciate and verify the diversity of scenarios of the CloudPeru2 dataset, we utilized t-SNE to sample the images by categories in a 2-D space, as shown in Fig.~\ref{fig:fig2}. For this, we used a small CNN of the same architecture as that of the network presented in \cite{cloudicann}, and we trained it in the SAT-6 airborne dataset \cite{deepsat}. Then, we resized the images of our dataset to $27\times27$ pixels and used the trained network to extract a vector of 128 features (i.e. the output of the first fully connected layer '$fc128$') from them. Finally, the extracted feature vectors are mapped to the 2D space with t-SNE.

\subsection{Proposed CNN for Segmentation}

We propose a semantic level segmentation of clouds in satellite imagery using a Convolutional Neural Network (CNN). The architecture of our network is the same as that used in \cite{mauritia} with the only difference that instead of three channels, our network uses inputs of four channels (R, G, B, and NIR). This CNN is based on the Deeplab v3+ architecture \cite{deeplabv3}, which integrates an encoder, an atrous spatial pyramid pooling module (ASPP), and a decoder.

\begin{figure} [!t]
\centering
\includegraphics[width=\columnwidth]{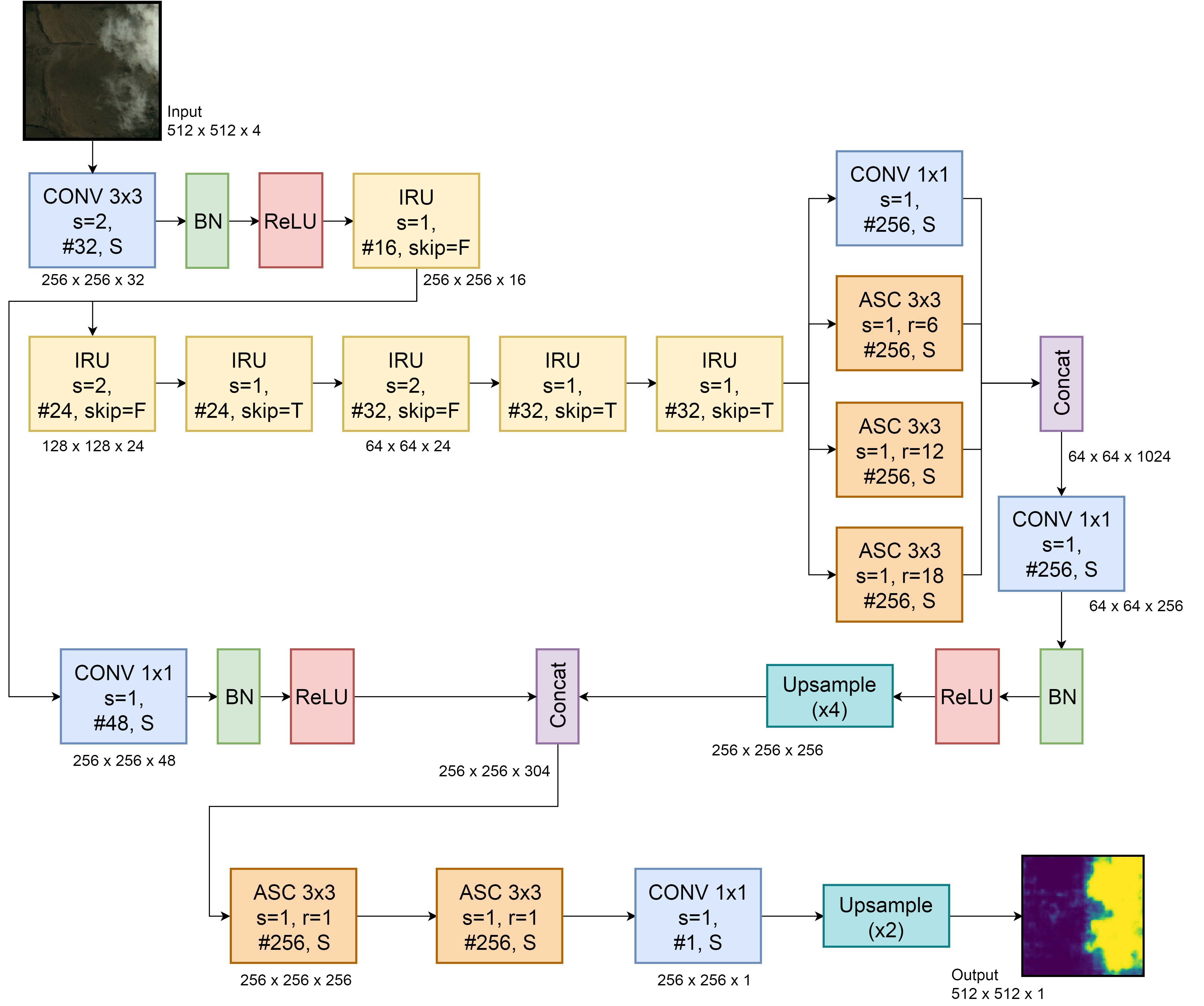}
\caption{The proposed network architecture. It uses regular convolutions ("CONV"), inverted residual units ("IRU") and atrous separable convolutions ("ASC").}
\label{fig:fig3}
\vspace{-2ex}
\end{figure}

In Fig.~\ref{fig:fig3} we show the proposed network architecture. Convolution blocks are denoted as  ``CONV;'' inverted residual units, as ``IRU;'' and atrous separable convolution blocks, as ``ASC''. The inverted residual unit (IRU) \cite{mobilenetv2} expands the input number of channels using a $1\times1$ convolution, then apply a $3\times3$ depthwise convolution (the number of channels remains the same), and, finally, apply another $1\times1$ convolution that reduces the number of channels. The atrous separable convolution (ASC) is a depthwise convolution with atrous convolutions followed by a pointwise convolution. The output number of filters of each block is reported using the hash symbol ("\#"). The stride of all convolutions is denoted as “s.” Blocks marked with “S” are “same padded,” which means that the output is the same size as the input. “ReLU” represents a standard rectified linear unit activation layer and “BN” a batch normalization layer.

\section{Results}

\subsection{Training Results and Metrics Comparison}

\begin{figure} [!b]
\centering
\includegraphics[width=\columnwidth]{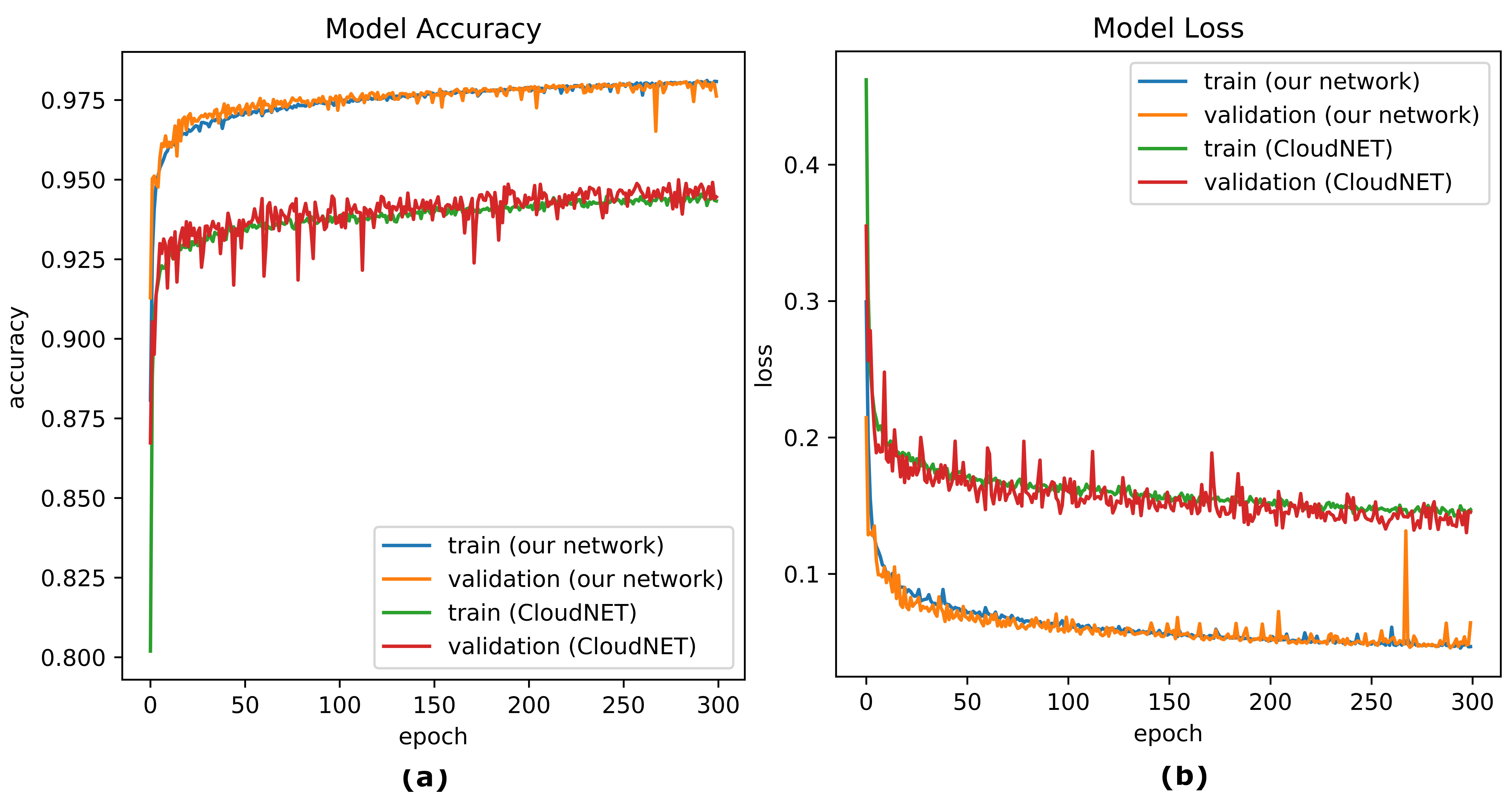}
\vspace{-4ex}
\caption{Training and validation results of our method and CloudNet \cite{cloudnet}. (a) Epochs vs. Accuracy (b) Epochs vs. Loss.}
\label{fig:fig4}
\vspace{-2ex}
\end{figure}

The proposed algorithm was implemented using Python 3.6 on a server with Intel Xeon CPU E5-2620 at 2.1 GHz CPU, 128GB RAM and two NVIDIA GeForce GTX 1080 GPU. The proposed CNN was trained using an Adam optimizer with a learning rate of 0.003, a momentum term \(\beta_1\) of 0.9, a momentum term \(\beta_2\) of 0.999 and a mini-batch size of 8. Figure~\ref{fig:fig4} shows the evolution of network accuracy and loss over 300 epochs.

\begin{figure*}[ht]
\centering
\includegraphics[height=10cm]{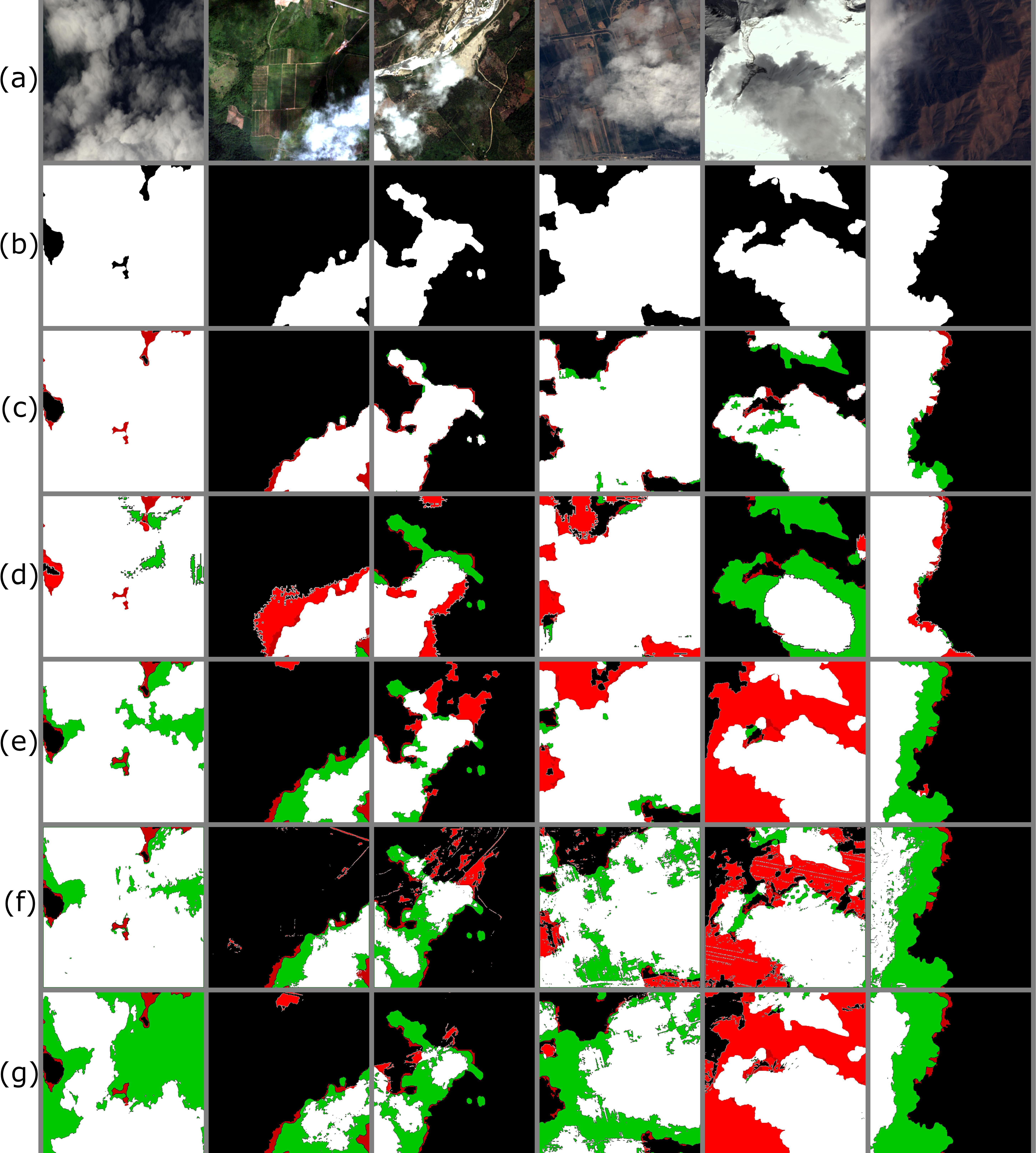}
\caption{Cloud segmentation using different methods. Green color represents False Negatives and red color, False Positives. (a) Original image (b) Ground truth (c) Our proposed method (d) CloudNet \cite{cloudnet}
(e) Method of \cite{cloudicann} (f) Method of \cite{cloudperuann} (g) Progressive refinement.}
\label{fig:fig5}
\vspace{-2ex}
\end{figure*}

In addition, we compared the ground truth with other four cloud detection methods. The first method \cite{cloudnet} proposes a new deep residual architecture called CloudNet to semantically segment clouds. Its main unit uses a $1\times1$ convolutional block of four channels followed by an ASPP module with seven dilation rates, whose results are concatenated along with the output of the first $1\times1$ convolution. By doing so, it preserves the spatial information since it does not use any pooling or strided operation. We implement a network with 12 of these units, according to best achieved results reported by the authors. The second method \cite{cloudicann} subdivides the image in superpixels, generates $27\times27$-pixel patches from each superpixel, and classifies each patch as cloud or non-cloud using a small CNN. The third method \cite{cloudperuann} calculates a set of texture and spectral descriptors and process them using a fully-connected neural network. Finally, the fourth method \cite{progressive} uses a progressive refinement scheme.

We quantitatively compare all methods with respect to the ground truth using five metrics in the validation set: accuracy (ACC), precision (PREC), recall/sensitivity (SN), and specificity (SP), as shown in Table~\ref{tab2}. The ACC ratio indicates the correctly predicted observations against total observations; the PREC ratio indicates the correctly predicted positive observations against the total predicted positive observations; the SN ratio indicates the correctly predicted positive observations against the total actual positive observations, and the SP ratio indicates the correctly predicted negative observations against the total actual negative observations. 

\setlength{\tabcolsep}{2pt}
\begin{table}[!b]
\caption{Metrics Comparison of Different Cloud Detection Methods}
\vspace{-4ex}
\begin{center}
\begin{tabular}{|c|c|c|c|c|c|}
\hline
\diagbox[width=23.5mm]{\textbf{Method}}{\textbf{Metric}} & \textbf{ACC (\%)} & \textbf{PREC (\%)} & \textbf{SN (\%)} & \textbf{SP (\%)} \\
\hline
\setlength\extrarowheight{9pt}

 Prog. ref.\cite{progressive} & 89.41 & 82.50  & 80.72  & 90.25 \\ 
\hline
 ANN \cite{cloudperuann} & 91.34 & 86.52 & 92.36 & 91.57 \\ 
\hline
 SLIC0 + CNN \cite{cloudicann} & 93.25 & 91.73 & 90.69 & 93.84 \\ 
\hline
 CloudNet \cite{cloudnet} & 94.01 & \textbf{97.82} & 89.78 & \textbf{98.04}   \\  
\hline
 Proposed method & \textbf{97.50} & 96.45 & \textbf{98.46} & 96.58 \\ 
\hline

\end{tabular}
\label{tab2}
\end{center}
\end{table}

From Table~\ref{tab2} we observe that the greatest accuracy and sensitivity values correspond to our method (97.5\% and 98.46\%), evidencing a difference of more than three and eight percentage points, respectively, over CloudNET. The visual comparison of the cloud masks generated by all mentioned methods is shown in  Fig.~\ref{fig:fig5}; these masks were generated from six different images with both low and high density clouds. It is observed that our method produces the most similar masks to the ground truth, specially when it comes to discern between snow and clouds (fifth column of Fig.~\ref{fig:fig5}), while other methods prioritize the segmentation of the most obvious high-density clouds. It is also worth mentioning that the most frequent type of error produced by our network is due to false positives, which can be proved by the fact that the lowest metrics of our network are the precision and specificity values; these errors are caused by small differences between the delineation of the borders of the ground truth and the generated masks. In the end, the results over the test set achieved an accuracy of 96.62\%, precision of 96.46\%, specificity of 98.53\%, and sensitivity of 96.72\%.

We would also like to state that although our version of CloudNet has only 6,077 trainable parameters and our method has 503,377, the amount of computation and memory required by our approach is inferior than that of CloudNet. For instance, when training CloudNet, we had to reduce the number of training samples to just 15,960, use a mini-batch size of 12, and use randomly cropped images of $200\times200$ in order to reduce the training time and memory consumption. This is explained by the fact that CloudNet does not reduce the size of its tensors at any moment, which consumes a lot of computational resources; while having small tensors with more number of channels consumes far less memory. Therefore, a single epoch for training our network (1330 batches, mini-batch size of 16, and inputs of $512\times512$ pixels) lasted 20 minutes, while for training CloudNet (1330 batches, mini-batch size of 12, and inputs of $200\times200$ pixels) lasted 22 minutes.
      
\subsection{Cloud Segmentation on Satellite Scenes}

\begin{figure} [!t]
\centering
\includegraphics[width=\columnwidth]{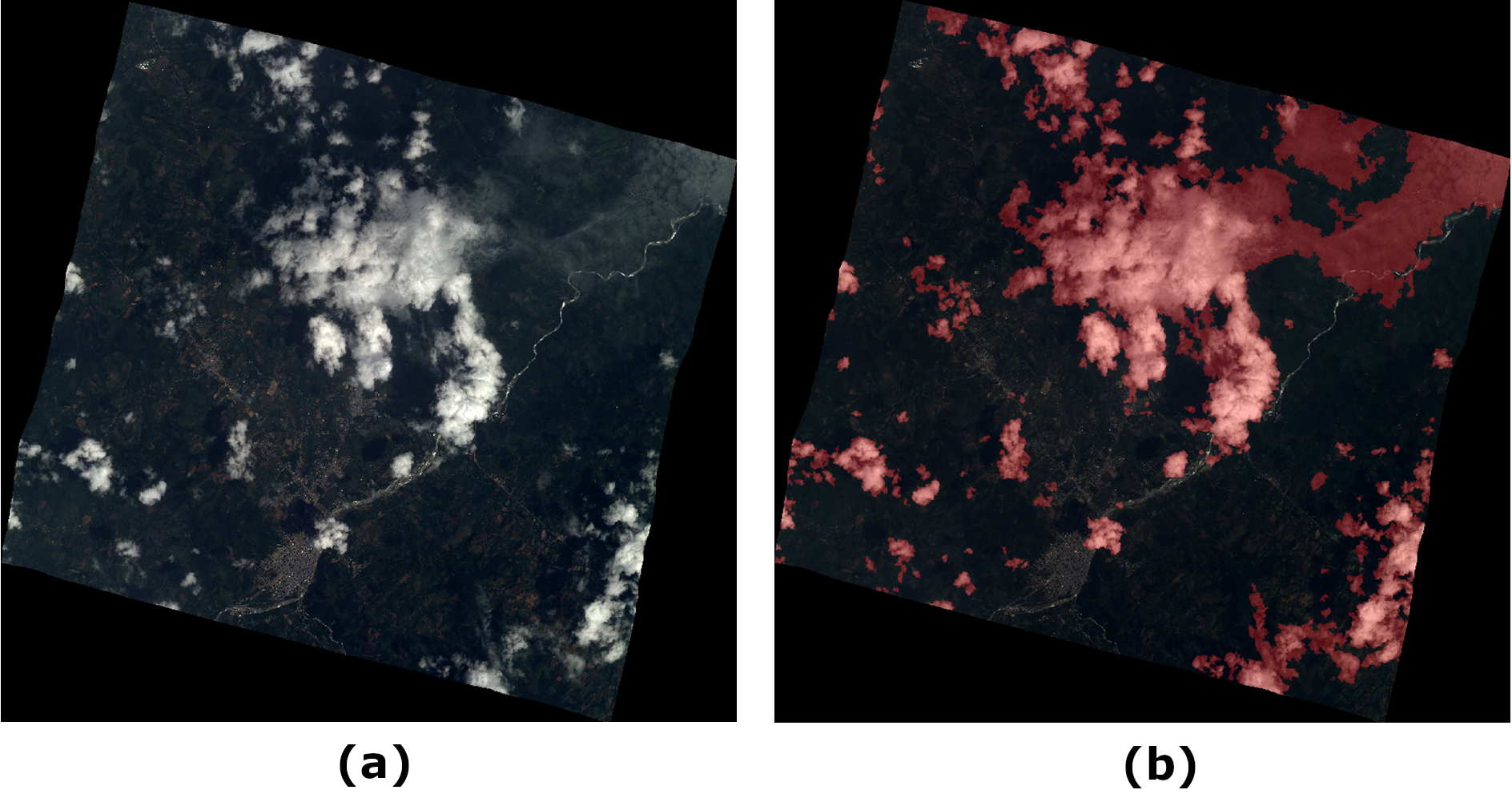}
\vspace{-4ex}
\caption{Cloud segmentation result in a satellite scene with clouds of many sizes and densities. (a) Original PERUSAT-1 scene in RGB. (b) Segmented clouds are painted in red over the original image.}
\label{fig:fig6}
\vspace{-2ex}
\end{figure}

We have trained a CNN to segment clouds on small patches of $512\times512$ pixels; however, the width and height of a PERUSAT-1 satellite scene are normally greater than 6000 pixels. Therefore, we move a $512\times 512$-pixel sliding window across the scene in both horizontal and vertical direction with a 50-pixel overlap. In each position, we get a cloud probability mask using the trained network. In the overlapped areas, we consider the maximum probability value in order to avoid discontinuities in the final mask. Finally, we apply a threshold of 0.5 over the entire mask. Figure~\ref{fig:fig6} shows the final cloud segmentation mask of a complete satellite scene.   

\section{Conclusions}

\balance
In this paper, we propose an efficient method for segmenting clouds in high-resolution multispectral satellite images semantically. For this, we trained an end-to-end convolutional neural network based on a simplification of Google\textquotesingle s Deeplab v3+ network. When comparing the results produced by our network with those produced by other cloud segmentation methods using the novel large dataset that we have proposed, we conclude that we achieved the best performance metrics. This method was embedded into a user-friendly interface used by the National Commission for Aerospace Research and Development (CONIDA) of Peru, allowing them to process hundreds of satellite images automatically and rapidly.
\section*{Acknowledgment}

The authors would like to thank the National Commission for Aerospace Research and Development (CONIDA) and the National Institute of Research and Training in Telecommunications of the National University of Engineering (INICTEL-UNI) for the support provided.

\end{document}